\documentclass[conference]{IEEEtran}
\IEEEoverridecommandlockouts
\usepackage{cite}
\usepackage{amsmath,amssymb,amsfonts}
\usepackage{algorithmic}
\usepackage{graphicx}
\usepackage{textcomp}
\usepackage{xcolor}
\def\BibTeX{{\rm B\kern-.05em{\sc i\kern-.025em b}\kern-.08em
    T\kern-.1667em\lower.7ex\hbox{E}\kern-.125emX}}

\usepackage{graphicx}
\usepackage{amsmath}
\usepackage{paralist}
\usepackage{framed}
\usepackage{amsfonts}
\usepackage{multirow}
\usepackage{courier}
\usepackage{subcaption}
\usepackage{array}

\usepackage[ruled,commentsnumbered,linesnumbered ]{algorithm2e}

\begin{document}

\title{Competitive Multi-Agent Deep Reinforcement Learning with Counterfactual Thinking}

\author{

\IEEEauthorblockN{Yue Wang${}^{1,2}$, Yao Wan${}^3$, Chenwei Zhang${}^4\IEEEauthorrefmark{1}$\thanks{\IEEEauthorrefmark{1}Work done while at University of Illinois at Chicago}, Lixin Cui${}^{1,2}$, Lu Bai${}^{1,2}$\IEEEauthorrefmark{2}\thanks{\IEEEauthorrefmark{2}Lu Bai is corresponding author (email: bailucs@cufe.edu.cn).}, Philip S Yu${}^{5}$}

\IEEEauthorblockA{${}^{1}$School of Information,
Central University of Finance and Economics, Beijing, P. R. China\\
Email: \{wangyuecs,bailucs,cuilixin\}@cufe.edu.cn}
\IEEEauthorblockA{${}^{2}$State Key Laboratory of Cognitive Intelligence, iFLYTEK, Hefei, P. R. China}
\IEEEauthorblockA{${}^{3}$Department of Computer Science, Zhejiang University, Hangzhou, P. R. China\\
Email: wanyao@zju.edu.cn}
\IEEEauthorblockA{${}^{4}$Amazon, Seattle, USA\\
Email: cwzhang@amazon.com}
\IEEEauthorblockA{${}^{5}$Department of Computer Science, University of Illinois at Chicago, Chicago, USA\\
Email: psyu@uic.edu}

}

\maketitle
%

\maketitle

\begin{abstract}
Counterfactual thinking describes a psychological phenomenon that people re-infer the possible results with different solutions about things that have already happened. It helps people to gain more experience from mistakes and thus to perform better in similar future tasks. This paper investigates the counterfactual thinking for agents to find optimal decision-making strategies in multi-agent reinforcement learning environments. In particular, we propose a multi-agent deep reinforcement learning model with a structure which mimics the human-psychological counterfactual thinking process to improve the competitive abilities for agents. To this end, our model generates several possible actions (intent actions) with a parallel policy structure and estimates the rewards and regrets for these intent actions based on its current understanding of the environment. Our model incorporates a scenario-based framework to link the estimated regrets with its inner policies. During the iterations, our model updates the parallel policies and the corresponding scenario-based regrets for agents simultaneously. To verify the effectiveness of our proposed model, we conduct extensive experiments on two different environments with real-world applications. Experimental results show that counterfactual thinking can actually benefit the agents to obtain more accumulative rewards from the environments with fair information by comparing to their opponents while keeping high performing efficiency.
\end{abstract}


\section{Introduction}

Discovering optimized policies for individuals in complex environments is a prevalent and important task in the real world. For example, (a) traders demand to explore competitive pricing strategies in order to get maximum revenue from markets when competing with other traders \cite{KUTSCHINSKI20032207};  (b) network switches need optimized switching logic to improve their communication efficiency with limited bandwidth by considering other switches \cite{10.1007/978-3-540-25947-3_4}; (c) self-driving cars require reasonable and robust driving controls in complex traffic environments with other cars \cite{DBLP:journals/corr/SallabAPY17}.

The core challenge raised in aforementioned scenarios is to find the optimized action policies for AI agents with limited knowledge about environments. Currently, many existing works learn the policies via the process of ``exploration-exploitation'' \cite{DBLP:conf/ewrl/CastronovoMFE12} which exploits optimized actions from the known environment as well as explores more potential actions on the unknown environment. From the perspective of data mining, this ``exploration-exploitation'' process can be considered as discovering the ``action-state-reward'' patterns that maximize total rewards from a huge exploring-dataset generated by agents.

There is a more complex situation that the environment may consist of multiple agents and each of them needs to compete with the others. In this scenario, it is imperative for each agent to find optimal action strategies in order to get more rewards than its competitors. An intuitive solution is to model this process as a Markov decision process (MDP) \cite{doi:10.1287/moor.12.3.441} and try to approach the problem by single-agent reinforcement learning, without considering the actions of other agents \cite{DBLP:journals/jair/KaelblingLM96}.
\begin{figure}[t!]
	\centering
	   \includegraphics[width=3.4in]{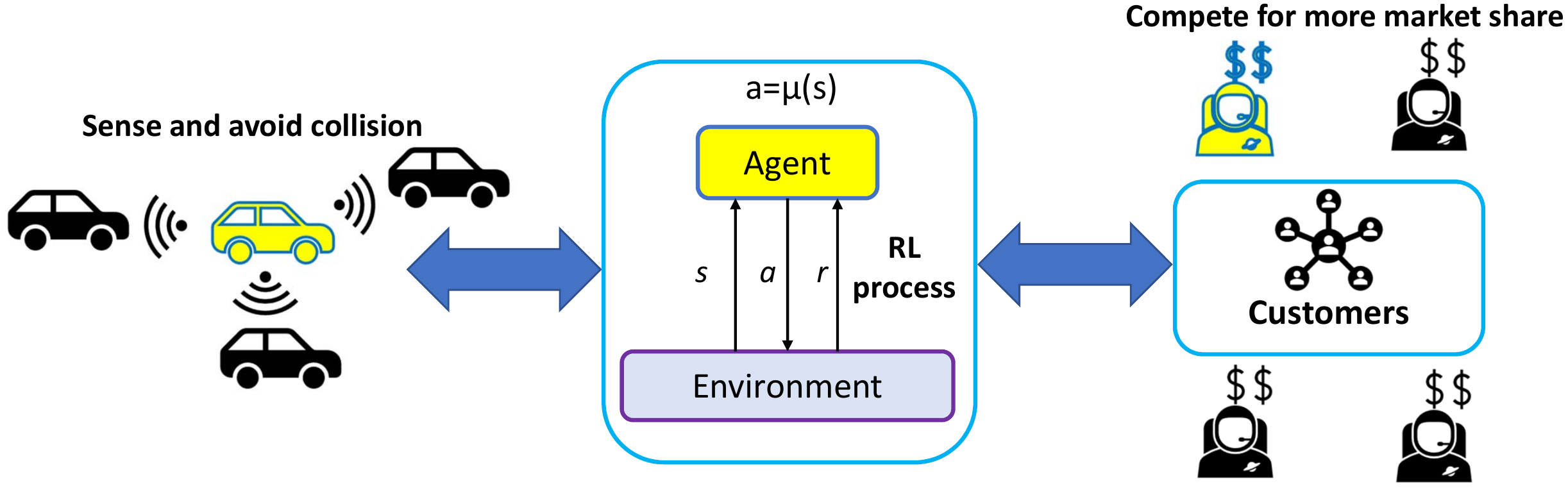}
	\caption{Explore and exploit the environments as RL processes.}
	\label{fig:rl_process} 
\end{figure}

Figure \ref{fig:rl_process} shows a schema of the process of reinforcement learning on two specific tasks, i.e., self-driving and marketing. Reinforcement learning aims to train agents to find policies which lead them to solve the tasks they do not have complete prior knowledge. Under the RL framework, an agent policy (i.e., $\mu(s)$ in Figure \ref{fig:rl_process}) is a probabilistic distribution of actions for the agent which is related to its observation or state for an environment. When an agent (i.e., car or trader) observes a new environment state, it performs an action and obtains a reward.

The RL training for agents is a greedy iterative process and it usually starts with a randomized exploration which is implemented by initializing a totally stochastic policy and then revising the policy by the received rewards at each iteration. The RL explores the policy space and favors those policies that better approximate to the globally optimal policy. Therefore, theoretically, by accumulatively exploring more policy subspaces at each iteration, the probability to get a better policy of an agent is increasing.

\noindent\textbf{\textit{Challenges.}} However, the traditional single-agent reinforcement learning approaches ignore the interactions and the decision strategies of other competitors. There are mainly two challenges to extend the reinforcement learning from single-agent to multi-agent scenarios. (a) \textit{Optimize action policy among competitors.} Generally, the single-agent reinforcement learning method (SRL) only optimizes the action policy for a specific agent. SRL does not model the interactions between multiple agents. Consequently, it challenges a lot when using SRL to optimize the action policy for a specific agent among a group of competitors simultaneously. (b) \textit{Learn action policy based on sparse feedbacks.} Since history never repeats itself, historical data only record feedbacks sparsely under the actions which have already happened, it challenges a lot to effectively learn optimized policies from historical data with sparse ``action-state-reward'' tuples. (c) \textit{Infer the counterfactual feedbacks.} One solution to the sparse feedbacks issue is to infer the counterfactual feedbacks for those historical non-chosen optional actions which have the potential to improve the learning efficiency for agent action policies. However, it still remains a challenge to counterfactually infer the possible feedbacks from an environment when an agent performs different optional actions at the same historical moment.

Currently, many existing works have applied the multi-agent deep reinforcement learning framework to mitigate the issues in environments with several agents. However, most of them \cite{DBLP:conf/nips/FoersterAFW16} \cite{DBLP:conf/nips/LoweWTHAM17} \cite{DBLP:conf/aaai/FoersterFANW18} still do not incorporate the counterfactual information contained in the history observation data which could further improve the learning efficiency for agents.

\noindent\textbf{\textit{Our Solutions and Contributions.}} To address the aforementioned challenges, in this paper, we formalize our problem as the competitive multi-agent deep reinforcement learning with a centralized critic \cite{DBLP:conf/nips/LoweWTHAM17} and improve the learning efficiency of agents by estimating the possible rewards for agents based on the historical observation. To this end, we propose a \underline{C}ounter\underline{F}actual \underline{T}hinking agent (CFT) with the off-policy actor-critic framework by mimicking the human-psychological activities.
The CFT agent works in the following process: when it observes a new environment state, it uses several parallel policies to develop action options or intents, for agents and estimates returns for the intents by its current understanding for the environment through regrets created by previous iterations. This is a similar process as the psychological activity that people reactive choices resulting from one's own experience and environment \cite{DBLP:conf/icml/ForneyPB17}. With the estimated returns, the CFT agent chooses one of the policies to generate its practical actions and receives new regrets for those non-chosen policies by measuring the loss between the estimated returns and practical rewards. This also mimics the human-psychological activities that people suffer regrets after making decisions by observing the gap between ideal and reality. Then the received regrets help the CFT agent to choose the policies in the next iteration.

It is worth mentioning that the proposed CFT agent is also more effective than existing multi-agent deep reinforcement learning methods during the training process since the parallel policy structure helps CFT agents to search from a wider range of policy subspaces at each iteration. Therefore, it could also be more informative than other related methods in multi-agent environments. We apply the CFT agent to several competitive multi-agent reinforcement learning tasks (waterworld, pursuit-evasion \cite{DBLP:conf/atal/GuptaEK17}) and real-world applications. The experimental result shows the CFT agents could learn more competitive action policy than other alternatives.

In summary, the main contributions of this paper can be summarized as follows:
\begin{itemize}
	\item We study the problem of competing with each other in a multi-agent environment with a competitive multi-agent deep reinforcement learning framework. Within this framework, we define the competitive ability of an agent as the ability to explore more policy subspaces.
	\item We propose the counterfactual thinking agent (CFT) to enhance the competitive ability of agents in multi-agent environments. CFT generates several potential intents through parallel policy structure and learns the corresponding regrets through the difference between estimated returns and practical rewards. The intent generation and regret learning process supervise each other with a max-min process. 
	\item We demonstrate that CFT agents are more effective than their opponents both in simulated and real-world environments while keeping high performing efficiency. This shows that counterfactual thinking mechanism helps agents explore more policy subspaces with the same iterations than other alternatives.
\end{itemize}
\noindent\textbf{\textit{Organization.}} The remainder of this paper is organized as follows.
In Section \ref{sec_preliminaries}, we introduce some background knowledge on multi-agent reinforcement learning. In Section \ref{sec_methodology}, we first give an overview of our proposed framework and then present each module of our proposed framework in detail. Section \ref{sec_experiments} describes the datasets and environments used in our experiment and shows the experimental results and analysis. Section \ref{sec_relatedwork} highlights some works related to this paper. Finally, we conclude this paper and give some future research directions in Section \ref{sec_conclusion}.
\section{Preliminaries}\label{sec_preliminaries}

When an agent tries to maximize its interests from an environment, it must consider both the reward it receives after each action and feedbacks from the environment. This could be simplified as a Markov decision process (MDP) and use reinforcement learning methods to search the optimal action policy \cite{vanOtterlo2012}. In our scenario, since our problem relates to the interaction between several agents, a natural way to explore this problem is to use an $N$-agents MDP \cite{DBLP:conf/atal/GuptaEK17}. In this section, we first introduce some background knowledge about multi-agent reinforcement learning and then mathematically formularized the problem studied in this paper.

\subsection{Multi-Agent Reinforcement Learning Framework}

In this paper, we consider a multi-agent extension of MDP named partially observable Markov games \cite{DBLP:conf/aaaifs/HausknechtS15}.
A Markov game for $N$ agents is defined by a set of states $\mathcal{S}$ describing the possible configurations of all agents, a set of actions, $\mathcal{A}_1, \mathcal{A}_2,\ldots,\mathcal{A}_N$, a set of states $\mathcal{S}$ which represents the observed environment settings of all agents and a state transition function $\mathcal{T}$.

\begin{equation}
\mathcal{T}: \mathcal{S}\times{\mathcal{A}_1}\times{\mathcal{A}_2}\times \ldots \times{\mathcal{A}_N}\mapsto{\mathcal{S}},
\end{equation}
where each agent $i  (i=1,2,3,\ldots,N)$ gets rewards as a function of the state and its action $r_i$:
\begin{equation}
r_i: \mathcal{S}\times{\mathcal{A}_i}\mapsto{\mathbb{R}}.
\end{equation}
The policy for each agent $i$ is a probabilistic distribution which is defined as:
\begin{equation}
\pi_i(a,s|\theta)\mapsto{[0,1]},
\end{equation}
where for $\forall{a}\in{\mathcal{A}_i}$ and $\forall{s}\in{\mathcal{S}}$. The target for each agent $i$ is to maximize its own accumulated expectation reward $R_{i}$:
\begin{equation}
R_{i}=\sum_{t=0}^{\infty}\gamma^{t}r^t_i,
\end{equation}
where $0<\gamma{<1}$ is a discount factor.

We adjust our setting by omitting the mapping from states to agent observations. This allows us to compare the competitive abilities of agents with different action policies under the same state information simultaneously.

\subsection{Temporal Difference Learning}
Many MDP problems are often solved by multi-agent reinforcement learning (MARL) \cite{DBLP:conf/nips/WangS02}. Since a real-world application contains more factors than agents could observe, we discuss our problem in the stochastic environment \cite{Lauer00analgorithm} with model-free methods. Monte Carlo and temporal difference learning (TD) are often used model-free methods to deal with reinforcement learning (RL) problems. Furthermore, since real-world applications are usually continuous without terminate states, we use TD methods to study the MARL problem in this paper.

The mainstream TD methods to learn the optimal policies for RL problems are categorized into value-based, policy-based \cite{BusoniuL2010} and combined methods (consider both the value and policy optimization). The representative methods include Q-learning \cite{DBLP:journals/ml/WatkinsD92}, policy gradient algorithms (PG)  \cite{DBLP:conf/nips/SuttonMSM99}, and actor-critic algorithms \cite{Kushner2003}. All these methods relates to two important notations: the value (V) function and action-value (Q) function  \cite{DBLP:conf/icml/SilverLHDWR14}.

If we let the agent optimize its policy independently,  the $V$ and $Q$ function for agent $i$ are denoted as follows.
\begin{equation}
V_i^{\pi}(s)=\mathbb{E}[R_i^1|s_1=s;\pi],
\end{equation}
\begin{equation}
Q_i^\pi(s,a)=\mathbb{E}[R_i^1|s_1=s,A_i^1=a;\pi],
\label{eq:original_q}
\end{equation}
where $R_i^t$ can be obtained by:
\begin{equation}
R_i^t=\gamma{^{0}}r_i(s_t,a)+\gamma{^{1}}r_i(s_{t+1},a)+\gamma{^{2}}r_i(s_{t+3},a)+....
\end{equation}
It is the total discounted reward from time-step $t$ for agent $i$. Intuitionally, $V_i^{\pi}(s)$ refers to the reward expectation of agent $i$ for state $s$ and $Q_i^\pi(s,a)$ represents the reward expectation of agent $i$ when it taking action $a$ at state $s$.

\subsection{Approximate Q and V with Deep Neural Networks}
In order to solve the combinatorial explosion problem \cite{DBLP:conf/ijcai/DutechBC01} in evaluating policies under high state or action dimensions, recent methods apply deep neural networks to estimate $V$ and $Q$ functions. This lead to the flourishing of deep reinforcement learning methods (DRL). The current popular DRL methods include Deep Q-Network (DQN) \cite{DBLP:journals/corr/MnihKSGAWR13} and Deep Deterministic Policy Gradient (DDPG) \cite{DBLP:journals/corr/LillicrapHPHETS15}.

DQN is the deep learning extension of Q-learning. It follows the value-based way and learns the Q-function with deep neural networks. Since DQN usually needs discrete candidate actions, and it may suffer non-stationary problems under multi-agent settings \cite{DBLP:conf/nips/LoweWTHAM17}, it is rare to use DQN in problems with continuous action spaces.

DDPG is a deep reinforcement learning method which combines a policy estimation and a value computation process together.  It originates from PG \cite{DBLP:conf/nips/SuttonMSM99} which models the performance of policy for agent $i$ as $J(\theta_i)=\mathbb{E}_{s\sim{\rho^{\pi_i}},a\sim\pi_i}(R_i)$. Then the gradient of the policy for agent $i$ is obtained by:
\begin{equation}
\nabla{J(\theta_i)}=\mathbb{E}_{s\sim{\rho^{\pi_i}},a\sim\pi_i^{\theta_i}}[\nabla_{\theta_i}\log\pi_i(a,s|{\theta_i})Q_i(s,a|\theta_i)],
\end{equation}
where  $\rho^{\pi_i}$  is the state distribution for agent $i$ by exploring the policy space with policy $\pi_i$.

Since computing PG requires to integrate over both state and action spaces, PG suffers from the high variance problem \cite{DBLP:conf/nips/LoweWTHAM17} and needs more samples for training. Deterministic policy gradient (DPG) \cite{DBLP:conf/icml/SilverLHDWR14} alleviates this problem by providing a continuous policy function $a=\mu(s|\theta_i)$ for agent $i$. This change avoids the integral over the action space.  With function $\mu(s|\theta_i)$, the gradient of DPG for agent $i$ can be written as:
\begin{equation}
\nabla{J(\mu_{\theta_i})}=\mathbb{E}_{s\sim{\rho^{\mu}}}[\nabla_{\theta_i}\mu(s|{\theta_i})\nabla_{a}Q_i(s,a|\theta_i)|_{a=\mu(s|{\theta_i})}].
\end{equation}

Since DPG only integrates over the state space, it can be estimated more efficiently than stochastic policy gradient algorithms \cite{DBLP:conf/icml/SilverLHDWR14} .

As a deep learning extension of DPG, by applying the off-policy actor-critic framework, DDPG \cite{DBLP:journals/corr/LillicrapHPHETS15} uses a stochastic behavior policy $\beta$ with noise in Gauss distribution to explore the state space $\rho^{\beta}$ and a deterministic target policy to approximate the critic policy. By learning the Q-values through neural networks, the gradient of agent $i$ in DDPG then becomes:
\begin{equation}
\nabla\!{J(\theta_i^{\mu})}\!=\!\mathbb{E}_{s_t\!\sim\!{\rho^{\beta_i}}}\![\nabla_{\theta^{\mu}_i}\!\mu\!(s_t|\theta^{\mu}_i)\!\nabla_{a}\!Q_i\!(s,a|\theta_i^{Q})|_{a\!=\!\mu(s_t|\theta^{\mu}_i)}],
\end{equation}
where $\theta_i^{Q}$ and $\theta_i^{\mu}$ are parameters for the target and current policy neural network respectively. During the training process, DDPG uses a replay buffer $D$ to record the ``state-action-reward'' tuples obtained by the exploration policy $\beta$ and then optimizes the parameters for the current neural network by drawing sample batches from $D$. With a trained current policy neural network, it updates the target policy neural network by a soft-updating method. This framework stabilizes the learning process and avoids the large variance problem in the original policy gradient methods and its deterministic action outputs are useful in continuous control RL problems. Therefore, DDPG has successfully applied in the MARL \cite{DBLP:conf/nips/LoweWTHAM17} problems and our model follows the similar off-policy actor-critic framework as DDPG.

\subsection{Compete in \textbf{\textit{N}}-Agent MDP}
In this paper, we aim to extend the reinforcement learning from single-agent to multi-agent settings in a competitive environment.
In order to make all agents compete with each other in an environment, we redefine the Q-values for all agents as the following equation.
\begin{equation}
Q_i^\pi\!(s_t,a) \!=\!{\gamma}Q_i^\pi\!(s_{t+1},a)\!+r'_i(s_t,a),(i=1,2,...,N),
\label{eq:revised_q}
\end{equation}
where $r'_i(s_t,a)$ is a revised rewards which is denoted as:
\begin{equation}
r'_i(s_t,a)={(1-\alpha)}r_i\!(s_t,a)\!+\!\alpha\frac{-r_{\hat{i}}(s_t,a)}{N-1}.
\label{eq:revised_r}
\end{equation}

In Equation \ref{eq:revised_r}, $r_{\hat{i}}(s_t,a)$ is the total rewards of all other agents than $i$; The weight $\alpha$ ($0\leq\alpha<1$) decides the ratio to consider the rewards of others for agent $i$. Therefore, the weight of $\alpha$ controls the degree of competition among all agents. e.g. when $\alpha>0.5$, the related agents care its own future rewards more than other the rewards of its other competitors.

With all agents maximizing the Q-values computed by Equation \ref{eq:revised_q}, a multi-agent environment becomes a more competitive environment than it is used to be.

\section{Counterfactual Thinking Agent in Multi-agent Reinforcement Learning}\label{sec_methodology}

\begin{figure*}[htb]
	\centering
	\begin{subfigure}[b]{2.3in}
		\includegraphics[width=\textwidth]{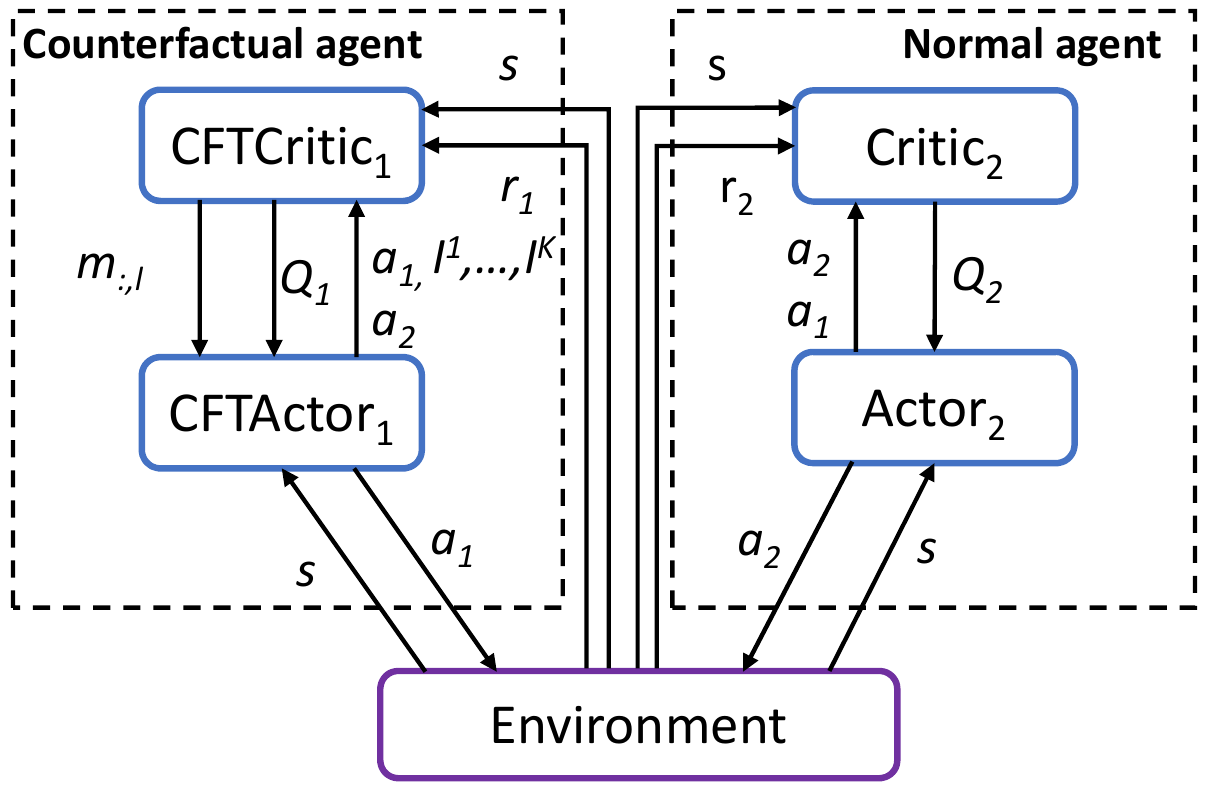}
		\caption{Framework.}
		\label{fig:subfig:a}
	\end{subfigure}
	\begin{subfigure}[b]{2.3in}
		\includegraphics[width=\textwidth]{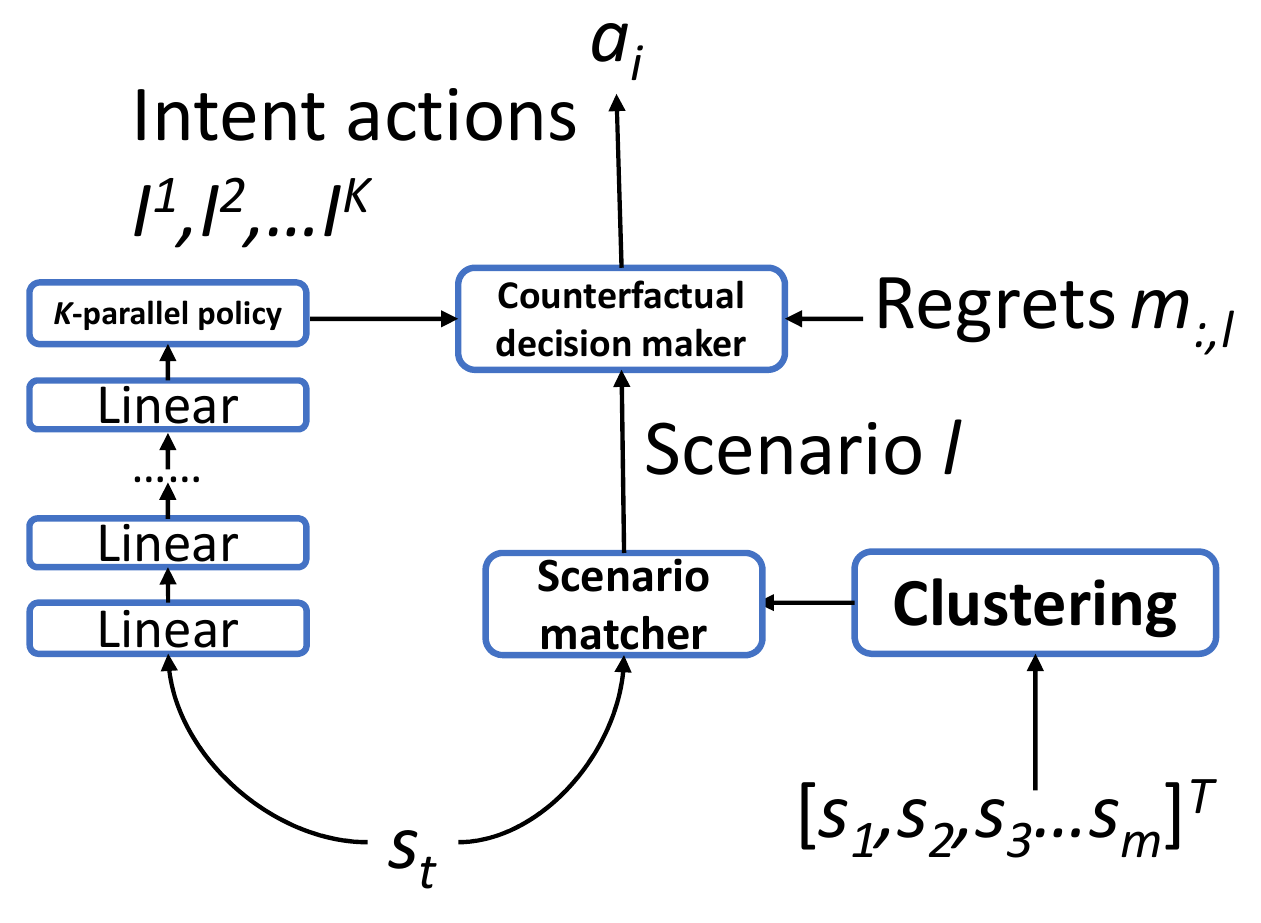}
		\caption{Counterfactual thinking actor.}
		\label{fig:subfig:b}
	\end{subfigure}
	\begin{subfigure}[b]{2.3in}
		\includegraphics[width=\textwidth]{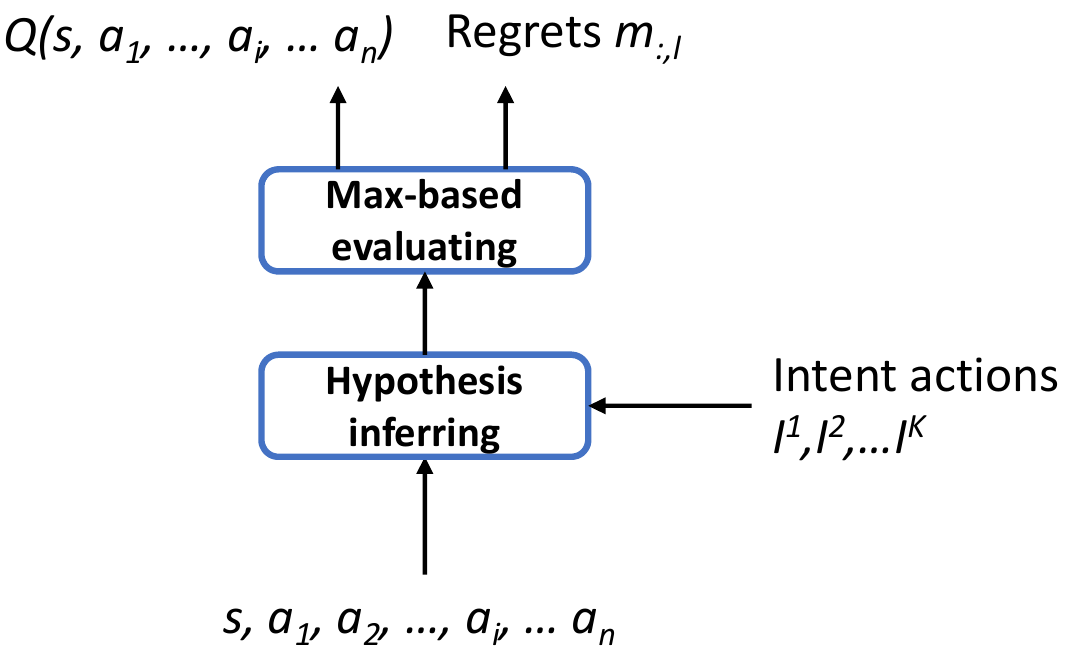}
		\caption{Counterfactual thinking critic.}
		\label{fig:subfig:c}
	\end{subfigure}
	\caption{The framework  counterfactual actor-critic reinforcement learning, where $a_i$ is the action of the $i$-th agent (the $i$-th agent is a counterfactual thinking agent) and $m_{:,l}$ is the regret distribution for the $l$-th scenario.}
	\label{fig_framework} 
\end{figure*}


Inspired by a psychological phenomenon, named counterfactual thinking, that people re-infer the possible results with different solutions about something that has already happened, this paper proposes to introduce a counterfactual thinking mechanism for an agent in a multi-agent environment. We argue that it may help people to gain more experience from mistakes and thus to perform better in similar future tasks \cite{Beldarrain2005}.
\subsection{An Overview}
As shown in Figure \ref{fig_framework}, our counterfactual thinking agent consists of a counterfactual thinking actor and a counterfactual thinking critic. Different from the previous actor-critic works \cite{DBLP:conf/nips/LoweWTHAM17}, our CFT agent generates several candidate actions by following its parallel policies when encountering a new state. Then, it chooses one of the candidate actions with its experience (historical regrets for parallel policies) to the environment. This mimics the behavior when a human seeking additional options before making decisions \cite{1688199}. Finally, the CFT agent revises its polices and the related regrets by evaluating the candidate actions with its current critic simultaneously. The experiment results show that this mechanism could bring more accumulative reward within the same episode for agents and thus make the related agents more competitive than their competitors. We will elaborate each module in detail in the following subsections.

\subsection{Counterfactual Thinking Actor}
A counterfactual thinking actor runs in the following process. It first matches the current encountered state into a specific scenario which belongs to the results of a state clustering process. Then it develops a few intents from $K$ parallel policies according to the matched scenario. Finally, it outputs both the minimum-regret action and all candidate actions to the environment and critic. With the Q-value estimated by the critic, the actor updates its regrets about the candidate policies. Concretely, we define the related notations for a counterfactual thinking actor as the followings.

\noindent\textbf{\textit{Counterfactual Thinking Actor.}} A counterfactual thinking (CFT) actor has $K$ parallel policies $\mu_i$ ($i=1,2,...,K$), where each policy associates to a regret $m_{k,l}$ ($m_{k,l}\in[0,1]$) under the $l$-th scenario. A counterfactual thinking actor generates $K$ intents $\mu_i(s)$.

To reduce the variance of the actions which are generated by counterfactual thinking actor, we combine a clustering process to the forward computation of the counterfactual thinking actor. As it is shown in Figure 2 (b), we use a clustering method to divide the encountered states of a counterfactual thinking actor to several clusters (named ``scenarios''). And then, every time our actor encounters a new state, it will check the clustering results to find the most related scenario and associates its policies with the corresponding regrets $\tau_i$s for that scenario.

In order to implement the parallel policy structure for counterfactual thinking actors, we propose the $K$-parallel policy layer in our actor which generates several intent actions with a given state.

\noindent\textbf{\textit{$K$-Parallel Policy Layer.}} A $K$-parallel policy layer contains a $\mathbb{R}^{K\times |g(s)|\times |a|}$ tensor $C=\{C^1,C^2,...,C^K\}$ where $C^k$ is a $\mathbb{R}^{|g(s)|\times |a|}$ matrix ($k=1, 2, ..., K$). The input for a $K$-parallel policy layer is a $\mathbb{R}^{|g(s)|}$ vector which represents an observed state. Its output includes $K$ vectors which represents $K$ intent actions or linear transformation of the intent actions. They can be computed as follows:
\begin{equation}
I^k=g(s)\times{C^k}, \mbox{ for } k=0,1,2,3,\ldots,K.
\end{equation}
$K$-parallel policy layer generates $K$ intent actions where $I^k$ is the $k$-th intent action; $s$ is the current state; $g(s)$ is a function which can be extended to several linear transformations or other neural layers.
After obtaining $K$ intent actions, we utilize a data structure named scenario-regret matrix to evaluate the parallel policies.

\noindent\textbf{\textit{Scenario-Regret Matrix.}} A scenario-regret matrix $M=\{m_{k,l}\}_{K\times L}$ is a $\mathbb{R}^{K\times L}$ matrix which records the regret values for $K$ policies under $L$ different scenarios. $m_{k,l}$ refers to the prior regret value for the $k$-th policy under the $l$-th scenario.
We get the scenarios through the aforementioned clustering process toward all encountered states by the actor. During the forward computation of the counterfactual thinking agent, every time a new state is observed, the actor first matches the state to a scenario (with the scenario matcher in Figure 2 (b)) and then outputs the intent with the minimum regret as the final action. The scenario matcher can be implemented by any kind of similarity computation.

The scenario-regret matrix is randomly initialized at first and then learned by receiving the regrets updated by the critic. The final output of a counterfactual thinking actor can be obtained through Algorithm 1.
\begin{algorithm}[h]
	\DontPrintSemicolon \KwData{state $s$, random degree $\epsilon$}
	\KwResult{action $a$}
	\Begin {
		Initialize the scenario-regret matrix $M$ randomly.\\
		Generate a set of $K$ intent actions $I$ with a $K$-parallel policy layer.\\
		Match the state $s$ to the $l$-the scenario.\\
		With probability $\epsilon$:\\
		~~~\textbf{Output} one of intent in $I$ as $a$ with the probabilistic distribution of SOFTMIN($M$).\\
		With probability $(1-\epsilon)$:\\
		~~~\textbf{Output} $a=\sum_{k\in{[1,K]}}m_{k,l}I^k$.
	}
	\caption{Forward computing for a CFT actor}
\end{algorithm}
The SOFTMIN function \cite{DBLP:journals/tnn/Romdhane06} in Algorithm 1 is an opposite operation to SOFTMAX which gives the policy with the minimum regret with the biggest weight. The random degree $\epsilon$ controls the ratio of Algorithm 1 to generate an action based on random sampling. Line 5-8 is the implementation for the ``counterfactual decision maker'' in Figure 2 (b). The counterfactual thinking actor is trained by the objective function.
 \begin{equation}
 \arg\max_{\theta^\mu_i}{q_i},
 \end{equation}
 where $q_i$ is computed by the current critic neural network with the state $s$ and the outputted $a$ from the CFT actor. $\theta^\mu_i$ is the parameters for the neural network of a CFT actor. Intuitively, this process revises the parameter $\theta^\mu_i$ for CFT actor to get the maximized $q_i$ at each iteration.

\subsection{Counterfactual Thinking Critic}
The critic in our model has two simultaneous tasks during the forward process: compute the Q-value and update the scenario-regret matrix for the counterfactual thinking actor. We discuss how this is implemented in this section.

\noindent\textbf{\textit{Counterfactual thinking critic.}} A counterfactual thinking critic computes the Q-values for all $K$ intent actions generated by the counterfactual thinking actor. By computing the maximum Q-value for all $K$ actions, it calculates the regret value for each intent actions.
Since the counterfactual thinking critic is a centralized critic, it also uses actions of all agents to evaluate Q-values. To compute maximum Q-value by considering all intent actions for an agent, we define the following notation.

\noindent\textbf{\textit{Counterfactual Q-value.}} In a multi-agent Markov game, if the $i$-the agent applies the counterfactual thinking mechanism (which means it uses the counterfactual thinking actor and critic), $s$ is the current state. The counterfactual Q-value $q_k^i$ can be obtained by Equation \ref{eq:original_q} with the current Q network.
	\begin{equation}
	q^k_i=Q(s,a_1,a_2,...,I_i^k,...,a_N),
    \label{eq:counterfactual_q}
	\end{equation}
	where $a_1$, $a_2$, ..., $a_{i-1}$, $a_{i+1}$, ..., $a_N$ are the actions of other $N-1$ agents in a multi-agent environment at this iteration. In Equation \ref{eq:counterfactual_q}, the action for the $i$-th agent is replaced by every intent action of the $K$ intent actions obtained by its counterfactual thinking actor.

For each iteration, our critic outputs the maximum counterfactual Q-value $\max(q_i)$ of all $q^k$s ($k=1,2,...,K$) for the $i$-th agent. With $\max(q_i)$, the posterior regrets for the $i$-th agent under the $l$-th scenario are computed by the following equation.
\begin{equation}
m^*_{k,l}=\max(q_i)-q_i^k,
\label{eq:regrets}
\end{equation}
where $k=1,2,...,K$. Then the objective function for a counterfactual thinking critic of the $i$-th agent is:
\begin{equation}
\arg\min_{\theta^{q_i},m_{:,l}}(\lambda\frac{|q_i^{t-1}-q_i^t|^2}{n}+(1-\lambda)KL(m_{:,l},m^*_{:,l})),
\end{equation}
where $q_i^{t-1}$ is the current Q-value computed by Algorithm 2 and $q_i^{t}$ is the target Q-value which can be computed by Equation \ref{eq:revised_q}.
The KL function is the KL-divergence which compares the difference between the prior and posterior regret distribution $m_{:,l}$ and $m^*_{:,l}$ for all $K$ intent action of the $i$-th agent.

\begin{algorithm}[h]
	\DontPrintSemicolon \KwData{state $s$, practical action $a$, an intent actions set $I$}
	\KwResult{$q^i_{t-1}$ and $m^*_{:,l}$}
	\Begin {
		Compute the $q^i_{t-1}$ by the current Q-neural-network with $s$ and $a$.\\
		Compute the $q_i^k$ for each intent $I^k$ generated by CFT actor.\\
		Find the maximum Q-value $\max(q_i)$ of all $q_i^k$s.\\
		Compute regrets $m^*_{:,l}$ for all intents by Eq. (15) under the $l$-th scenario.\\
		~~~\textbf{Output} $q^i_{t-1}$ and $m^*_{:,l}$.
	}
	\caption{Forward computing for a CFT critic}
\end{algorithm}

In Algorithm 2, Line 3 corresponds to the ``Hypothesis inferring'' and Line 4-5 corresponds to the ``Max-based evaluating'' in Figure 2 (c) respectively.

\subsection{End-to-End Training}
Our CFT agent consists of a counterfactual thinking actor and a counterfactual thinking critic. Since both the forward processes of them are differentiable, we train this model with the back-propagation methods with an Adam \cite{DBLP:journals/corr/KingmaB14} optimizer. The training for CFT agents is a max-min process \cite{DBLP:conf/mlhc/ChoiBMDSS17} which maximizes the Q-value for the actor with current critic and minimizes the difference between the current and target critics. Since the CFT actor and critic are linked by a scenario-regret matrix, during the training process, the actions outputted by the CFT actor are weighted by the scenario-regret matrix learned by last iteration and the CFT critic revises the scenario-regret matrix with its forward process.

\section{Experiments and Analysis}\label{sec_experiments}
To verify the effectiveness of our proposed CFT, we conduct experiments on two standard multi-agent environments with real-world applications. Overall the empirical results demonstrate that the CFT excels on competitive multi-agent reinforcement learning, consistently outperforming all other approaches.

\subsection{Compared Baselines} 
The comparison methods of this work are MADDPG \cite{DBLP:conf/nips/LoweWTHAM17}, CMPG \cite{DBLP:conf/aaai/FoersterFANW18} and our counterfactual thinking agent (CFT).
\begin{itemize}
	\item \textbf{MADDPG} is the state-of-the-art method about the multi-agent deep reinforcement learning. Since our model is based on the similar off-policy actor-critic framework as MADDPG, the comparison of MADDPG and our model can directly tell us whether the proposed counterfactual mechanism improves the competitive ability for an agent.
	
	\item \textbf{CMPG} uses historical actions of agents as the estimated intents to enhance the stability of the learning process for the actor-critic framework. Since CMPG is the latest methods which improve the learning efficiency for RL problems with a counterfactual style method, we also compare our model with it.
	
\end{itemize}

\subsection{Environment 1: Multi-Agent Water-World (MAWW)}\label{sec_maww}
\begin{figure*}[!t]
	\centering
	\begin{subfigure}[b]{2.3in}
		\includegraphics[width=2.3in,height=1.7in]{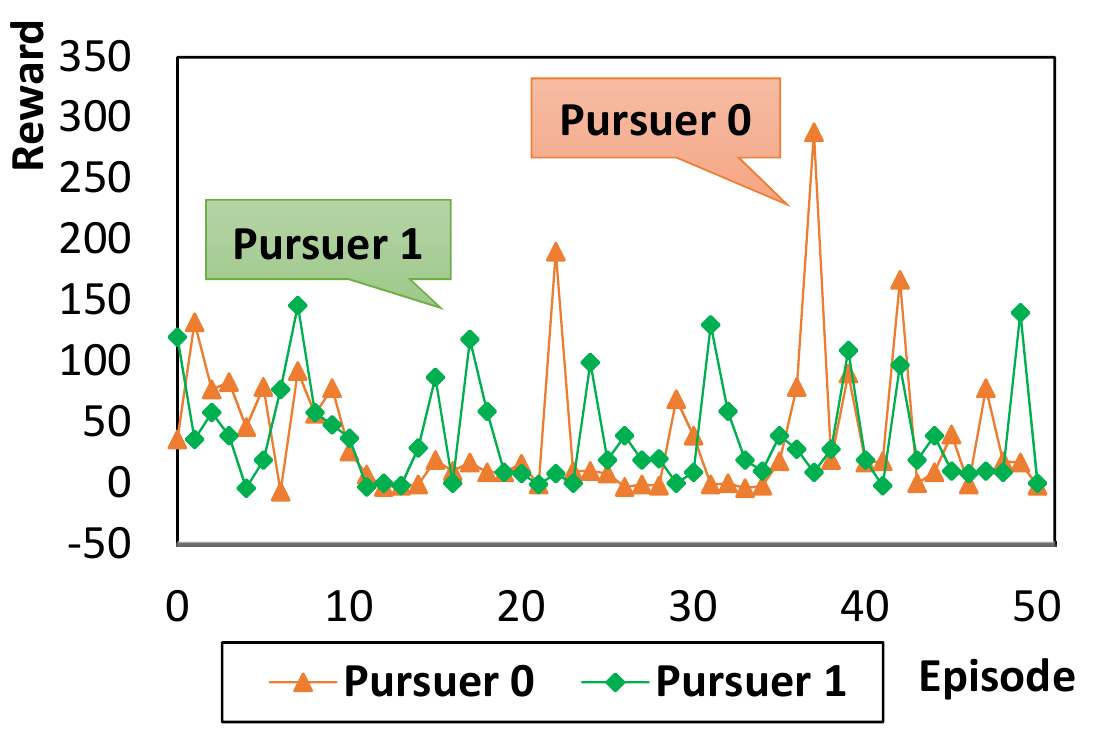}
		\caption{Both pursuer 0 and 1 are DDPG agents (MADDPG).}
		\label{fig:subfig:a}
	\end{subfigure}
	\begin{subfigure}[b]{2.3in}
		\includegraphics[width=2.3in,height=1.705in]{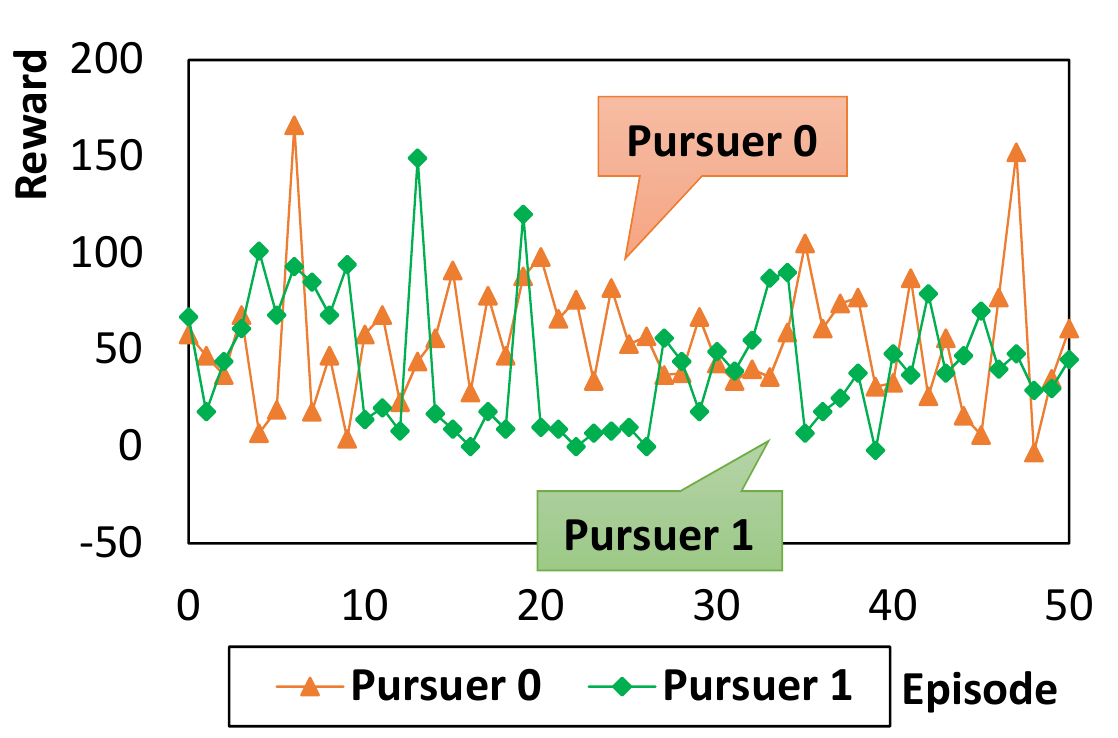}
		\caption{Pursuer 0 is a CFT agent, pursuer 1 is a DDPG agent.}
		\label{fig:subfig:b}
	\end{subfigure}
	\begin{subfigure}[b]{2.3in}
		\includegraphics[width=2.3in,height=1.7in]{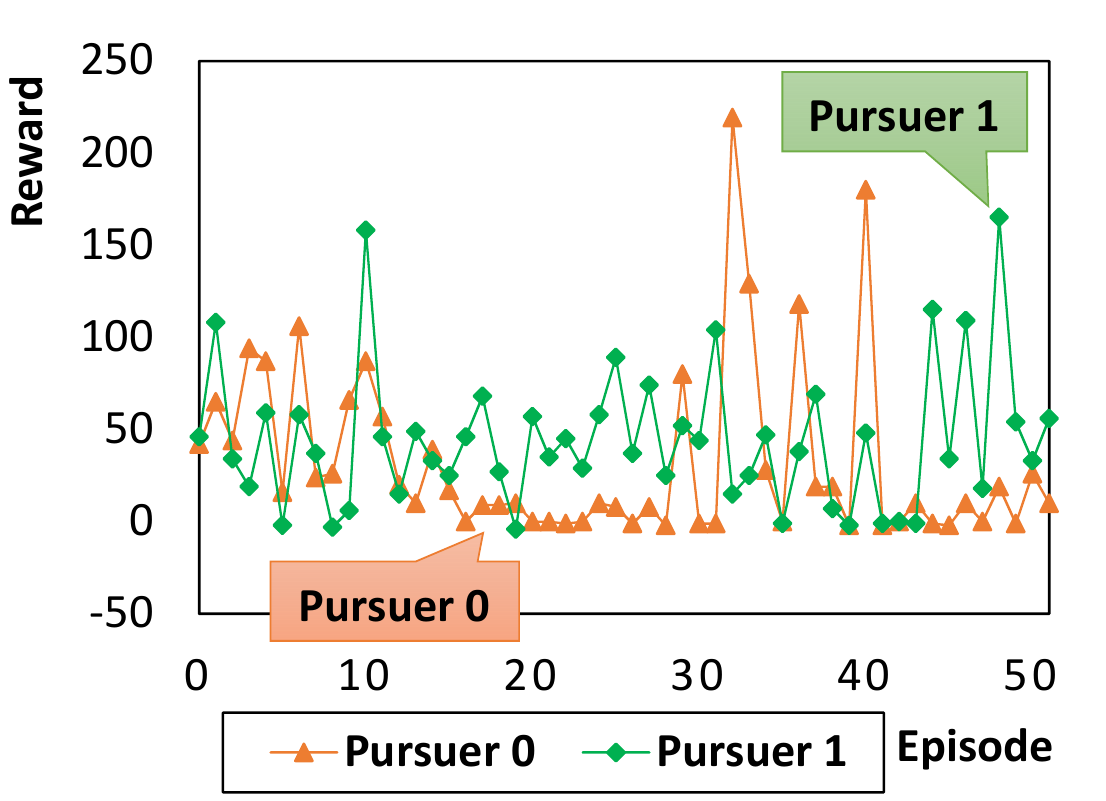}
		\caption{Pursuer 1 is a CFT agent, pursuer 0 is a DDPG agent.}
		\label{fig:subfig:c}
	\end{subfigure}
	\begin{subfigure}[b]{2.3in}
		\includegraphics[width=2.3in,height=1.7in]{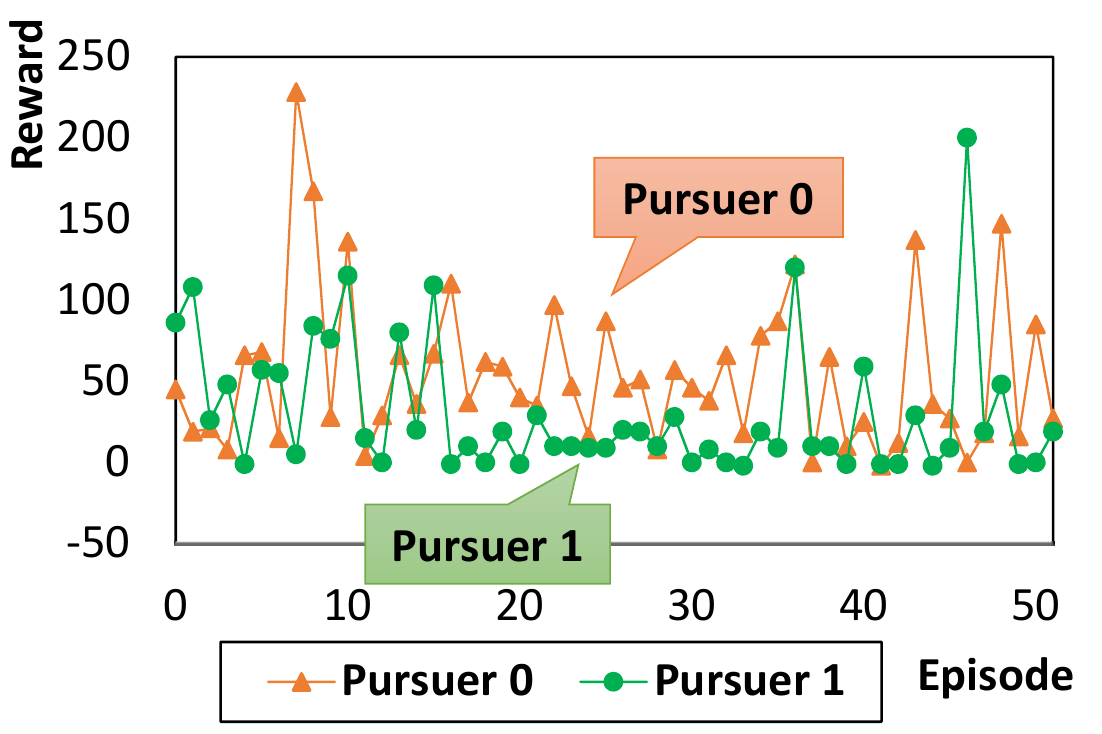}
		\caption{Pursuer 0 is a CMPG agent, pursuer 1 is a DDPG agent.}
		\label{fig:subfig:d}
	\end{subfigure}
	\begin{subfigure}[b]{2.3in}
		\includegraphics[width=2.3in,height=1.7in]{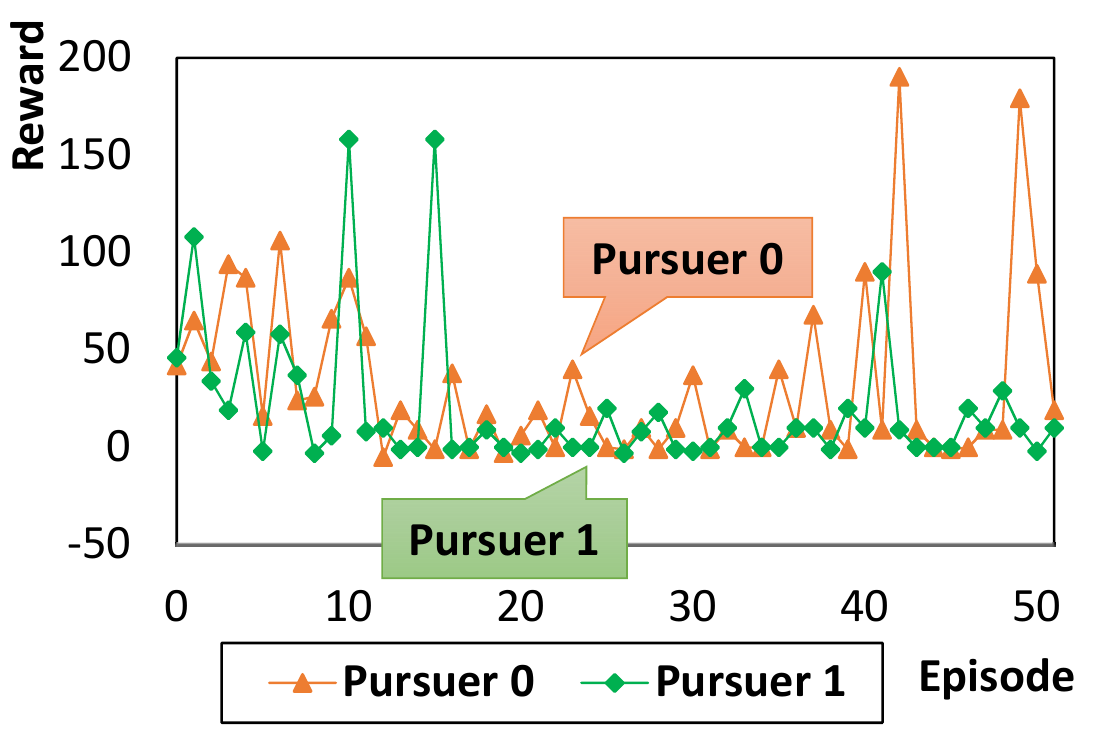}
		\caption{Pursuer 1 is a CMPG agent, pursuer 0 is a DDPG agent.}
		\label{fig:subfig:e}
	\end{subfigure}
	\begin{subfigure}[b]{2.3in}
		\includegraphics[width=2.3in,height=1.7in]{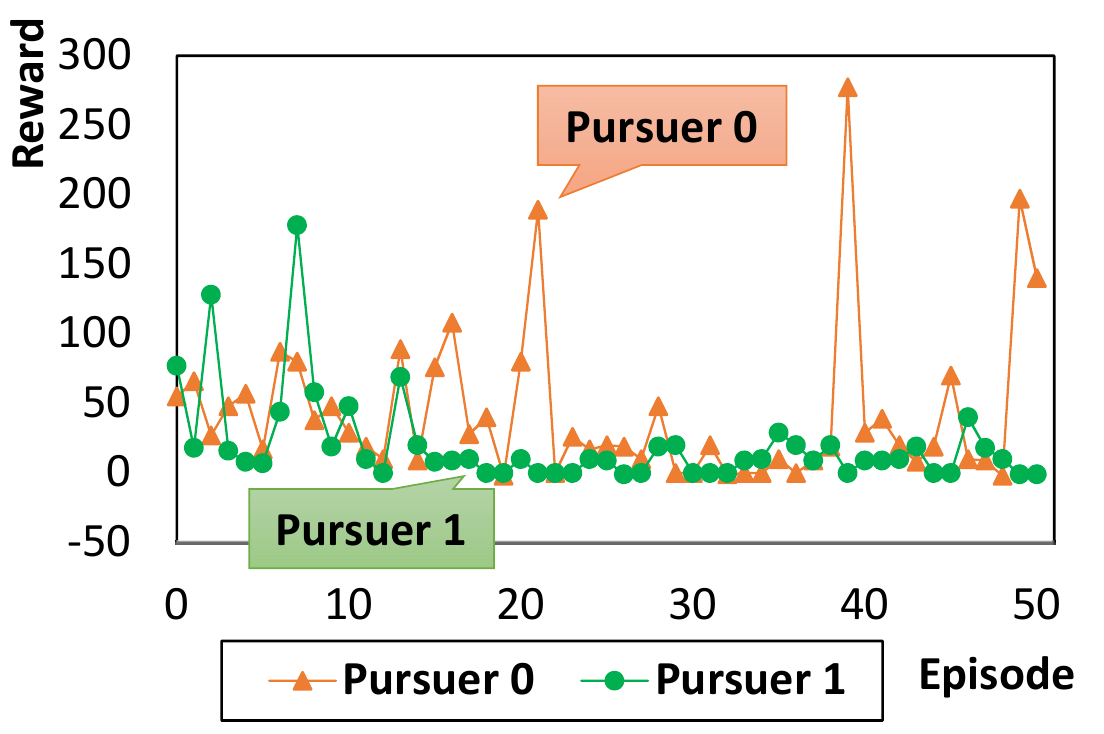}
		\caption{CFT V.S. CMPG. Pursuer 0 is a CFT agent and pursuer 1 is a CMPG agent.}
		\label{fig:subfig:f}
	\end{subfigure}
	\caption{Comparison of accumulative rewards obtained by agents on MAWW environment. (a) directly use the framework of MADDPG and it is shown no significant difference between the competitive abilities of two related agents; (b) and (c) compares the competitive abilities between a pre-set counterfactual agent and an agent applying DDPG. (d) and (e) shows that CMPG can also improve the competitive abilities for agents towards DDPG based competitors; (f) shows that a CFT agent is more competitive than a CMPG agent.}
	\label{fig:subfig}
\end{figure*}
\noindent\textbf{\textit{Problem Background.}} This is a multi-agent version pursuer and evader game in a simulated underwater environment which is provided in MADRL \cite{gupta2017cooperative}. Several pursuers are co-existing to purchase some evaders in an environment with floating poison objects and obstacles. Every time a pursuer captures an evader, it receives +10 reward. What's more, the pursuer receives -1 reward when it encounters a poisoned object. This environment can be used to research the mixed cooperative-competitive behaviors between agents.

\noindent\textbf{\textit{Training Setup.}}
We set the scenario number to 16 and the number of intent actions $K$ to 4. Furthermore, since our method is based on the off-policy actor-critic framework, we set the exploration episode to 10 for each testing below. This means that the policies for all agents are optimized after the $10^{th}$ episode.

\noindent\textbf{\textit{Results.}} In this experiment, we add two pursuers to compete in a water world environment with 50 evaders and 50 poison objects. To compare the competitive ability for each mentioned method, we set one of the pursuers as a CFT or CMPG agent, the others as DDPG agents. Furthermore, we also compare the competitive abilities of CFT and CMPG. As it is shown in Figure 3 (a), the rewards for two same DDPG agents are almost the same. This means that there is no difference between the competitive abilities of two DDPG agents. We further analyze the results in Figure 3 (b) and (c) and discover that the CFT agent receives significant more rewards than its DDPG based competitors. Figure 3 (d) and (e) present that CMPG can also improve the competitive abilities for an agent in this task. Figure 3 (f) compares the competitive abilities for CFT and CMPG agent directly, the result shows that CFT agent can be more competitive than the CMPG agent. In addition, we analyze the means and standard deviations of rewards in all cases in Figure 2. The results are listed in Table 2, where the last row (corresponding to Figure 3 (f)) compares the competitive abilities of a CFT agent (pursuer 0) and a CMPG agent (pursuer 1). The star-marked pursuers in Table 2 apply CFT or CMPG and the none-star-marked pursuers apply DDPG. In all, the results in this section confirm that our counterfactual mechanism (CFT) indeed helps an agent to compete in the multi-agent environments.

\subsection{Environment 2: Multi-Seller Marketing (MSM)}

\noindent\textbf{\textit{Problem Background.}} In the Multi-Seller Marketing (MSM) environment, a market contains multiple sellers is a perfect environment to fit the multi-agent Markov game framework. To study the dynamic process between sellers in a multi-seller market, we conduct experiments on two real-world datasets (i.e. RETAIL\footnote{https://www.kaggle.com/c/acquire-valued-shoppers-challenge/data} and HOTEL\footnote{https://www.kaggle.com/c/expedia-personalized-sort/data}) of the MSM environment. Table \ref{table_data_statistic} shows the statistics of these two datasets. We use the first 100,000 rows of RETAIL and all rows of HOTEL in this experiment.
\begin{table}[h]
	\centering
	\begin{tabular}{l|cc}
		\hline
		&RETAIL&HOTEL\\
		\hline
		Rows&100,000&276,592\\
		Start&2012-03-02&2012-11-01\\
		End&2013-07-23&2013-06-30\\
		Competitor num.&2,606&57,646\\
		\hline
	\end{tabular}
	\caption{Dataset statistics}
	\label{table_data_statistic}
\end{table}

Those datasets (i.e., RETAIL and THOTEL) contain the instant price as well as the volume of products (or hotel booking count), for different brands. We treat the instant prices as the actions and the corresponding sales volume as the reward for the corresponding brand sellers. Since our model needs a centralized critic, the state for each agent is the same with others which consists of instant sales volumes for all sellers. To predict the feedbacks of a market, we use a recurrent neural network model (RNN) \cite{DBLP:conf/nips/SutskeverVL14} to learn the relationship between the instant prices and rewards (by modeling the prediction as a sequence-to-sequence learning problem).

\noindent\textbf{\textit{Training Setup.}}
We set the scenario number to 16 and the number of intent actions $K$ to 6. Moreover, we also set the exploration episode to 10 for each testing as in Section \ref{sec_maww}.

\begin{table}[h]\small
	\centering
	\begin{tabular}{lccc}
		\hline
		Methods&MADDPG&CFT&CMPG\\
		\hline
		Pursuer0$^{*}$&38.2$\pm$57.13&\color{blue}\textbf{61.0$\pm$33.62}&54.2$\pm$47.32\\
		Pursuer1&37.9$\pm$42.04&\color{blue}\textbf{22.2$\pm$31.17}&48.3$\pm$47.51\\
		\hline
		Pursuer0&38.2$\pm$57.13&\color{blue}\textbf{32.9$\pm$48.39}&31.9$\pm$43.59\\
		Pursuer1$^{*}$&37.9$\pm$42.04&\color{blue}\textbf{44.3$\pm$38.31}&19.5$\pm$36.09\\
		\hline
		CFT vs CMPG&-&\color{blue}\textbf{43.4$\pm$55.83}&19.7$\pm$32.89\\
		\hline
	\end{tabular}
	\caption{Comparison of accumulative rewards. The star marked pursuer is applying the competitive models.}
\end{table}

\noindent\textbf{\textit{Results on RETAIL dataset in the MSM environment.}} Figure 4 lists the comparison results between CFT and DDPG agents on MSM with retail datasets. In this section, we extract the price timeseries and sales volumes for top-7 sellers of the best-seller product from RETAIL dataset. As introduced in 4.1, we trained RNN with the extracted results. Based on Figure 4 (a), we analyze whether the RNN prediction model captures the real-world rules. In order to ease the observation about the difference of seller behaviors with the different market occupation, we rank all sellers according to their accumulative sales volumes and named them as ``Seller 0'' to ``Seller 6''.

To further analyze the effective for counterfactual thinking mechanism, we let seller 3 or 6 as the CFT agents respectively. The result is shown in Figure 3 (b) and (c). We observe that the ranks of seller 3 and 6 are highly improved after using the CFT method in Figure 3 (b) and (c).

\noindent\textbf{\textit{Results on HOTEL dataset in the MSM environment.}} Figure 5 compares the competitive abilities of agents on MSM with HOTEL dataset. In this section, we extract the price timeseries and sales volumes for top-5 brands of the most popular hotels from HOTEL dataset. In a similar way of the last section, we rank all hotel brands according to their accumulative sales volumes and named them as ``Hotel 0'' to ``Hotel 9''.

In this testing phase, we let hotel 0 or 1 to learn policies with counterfactual mechanism respectively since they are the two least competitive agents in Figure 4 (a). From Figure 4 (b) and (c), we can observe that the accumulative rewards of a CFT agent are significantly increased under the same environment by competing with other DDPG based competitors (hotels).

In summary, both the results in Figure 3, 4 and 5 show that our counterfactual mechanism indeed helps an agent to become more competitive than it before in a multi-agent Markov game environment.

\begin{figure*}[!t]
	\centering
	\begin{subfigure}[b]{2.3in}
		\includegraphics[width=\textwidth]{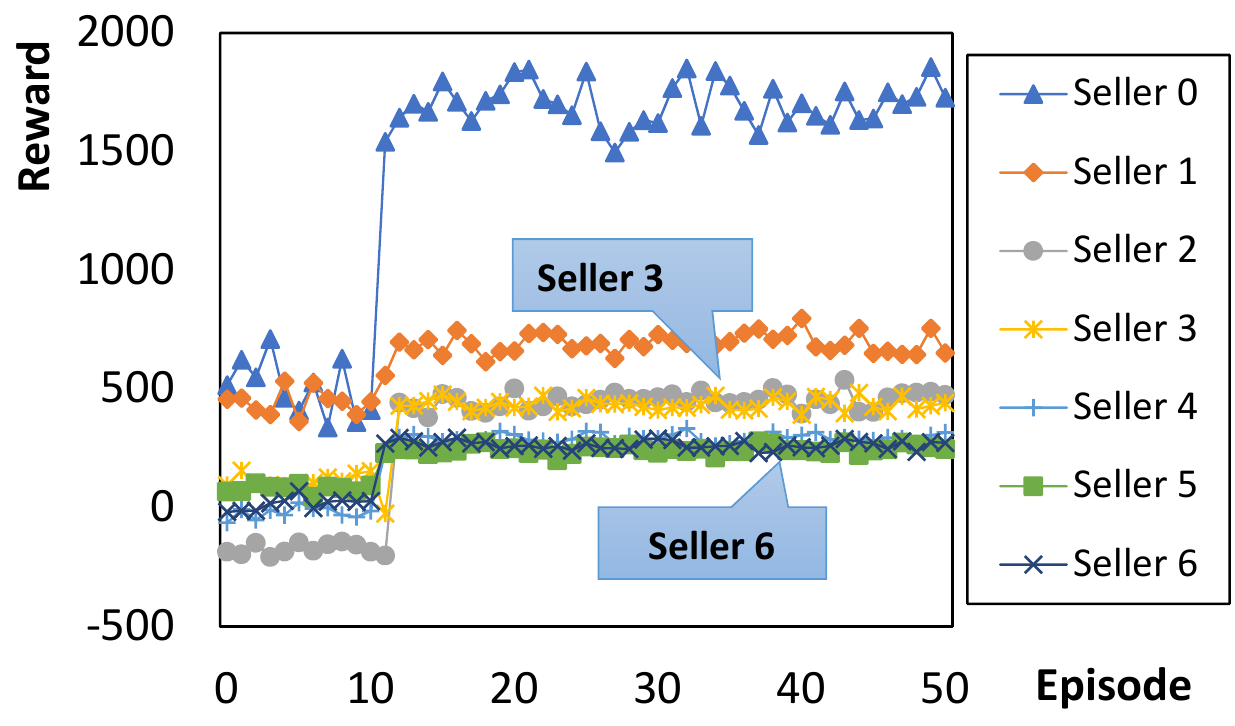}
		\caption{All sellers are DDPG agents, which is the default MADDPG.}
		\label{fig:subfig:a}
	\end{subfigure}
	\begin{subfigure}[b]{2.3in}
		\includegraphics[width=\textwidth]{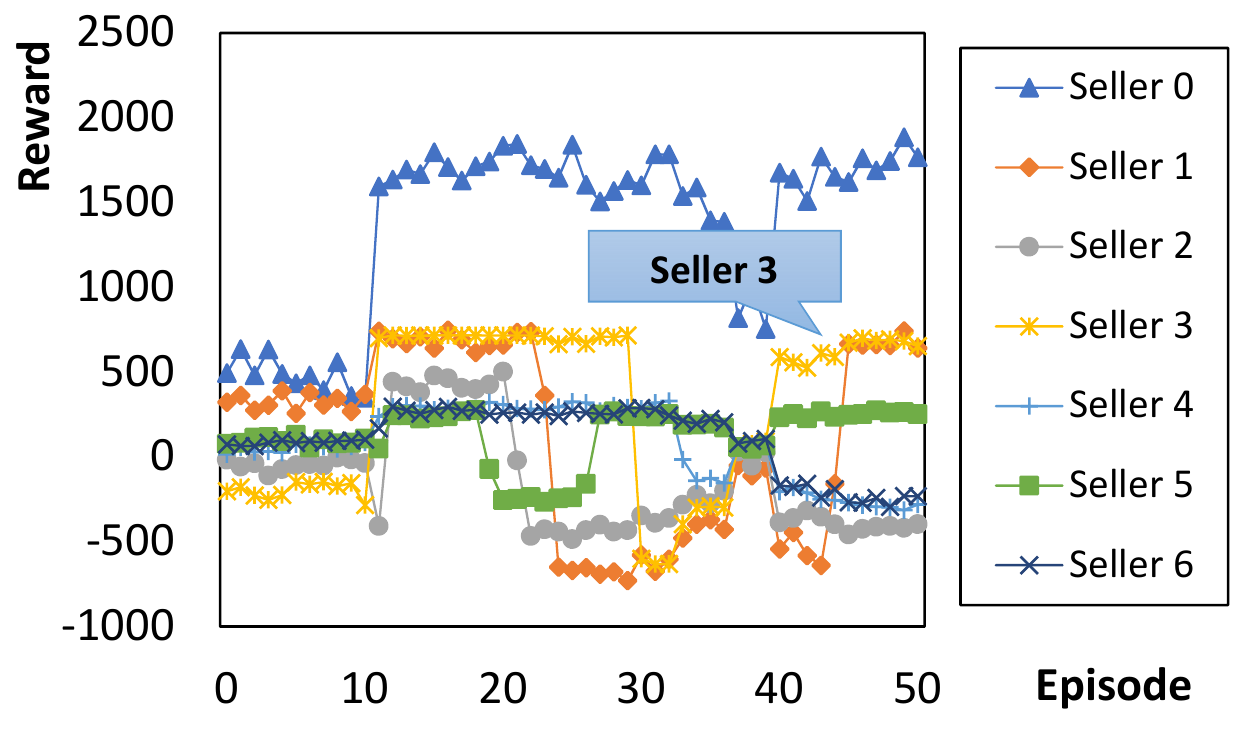}
		\caption{Seller 3 is a CFT agent, others are DDPG agents}
		\label{fig:subfig:b}
	\end{subfigure}
	\begin{subfigure}[b]{2.3in}
		\includegraphics[width=\textwidth]{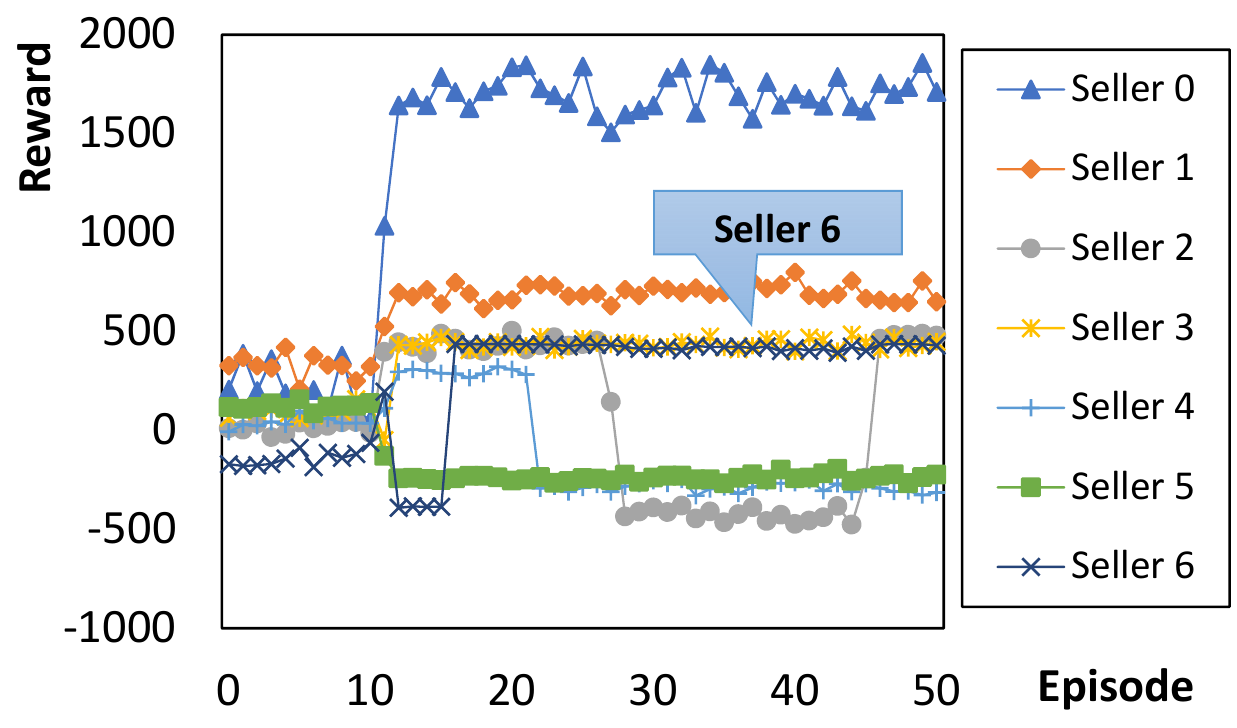}
		\caption{Seller 6 is a CFT agent, others are DDPG agents}
		\label{fig:subfig:c}
	\end{subfigure}
	\caption{Comparison of different actors thinking with counterfactual actor-critic reinforcement learning on MSM with RETAIL dataset.}
	\label{fig:subfig}
\end{figure*}

\begin{figure*}[!t]
	\centering
	\begin{subfigure}[b]{2.3in}
		\includegraphics[width=\textwidth]{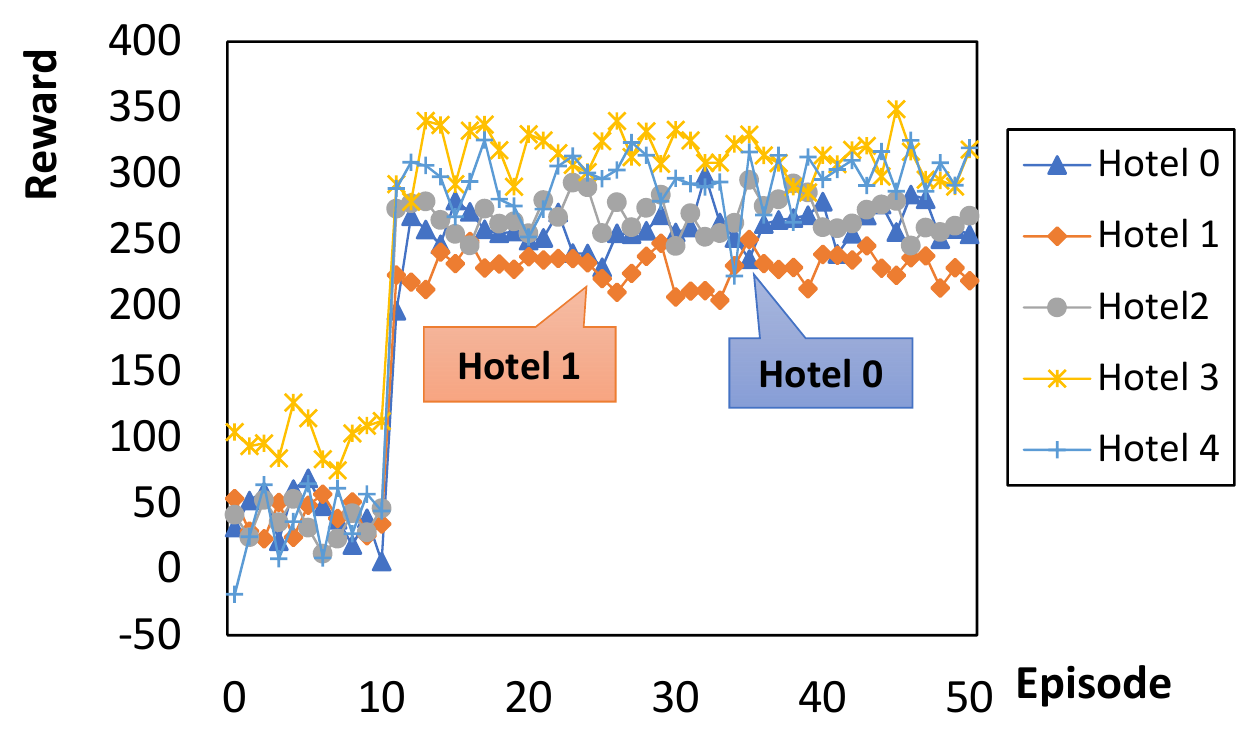}
		\caption{All hotels are DDPG agents, which is the default MADDPG.}
		\label{fig:subfig:a}
	\end{subfigure}
	\begin{subfigure}[b]{2.3in}
		\includegraphics[width=\textwidth]{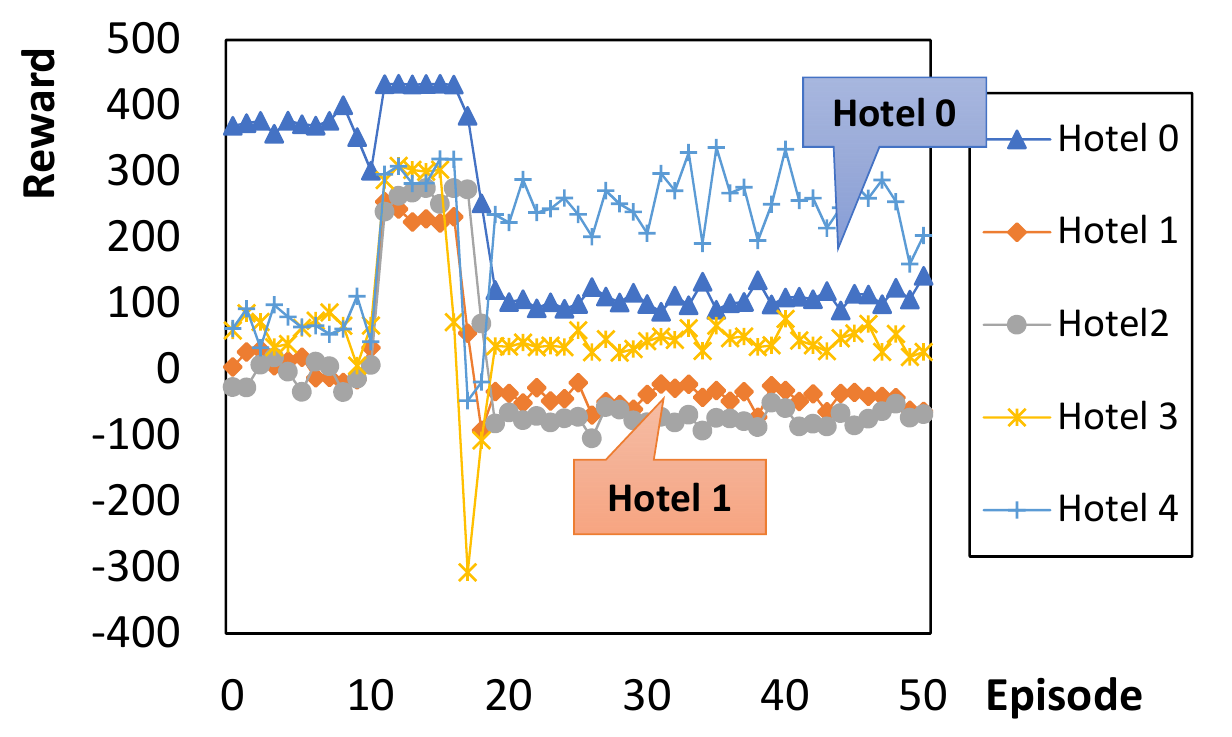}
		\caption{Hotel 0 is a CFT agent, others are DDPG agents}
		\label{fig:subfig:b}
	\end{subfigure}
	\begin{subfigure}[b]{2.3in}
		\includegraphics[width=\textwidth]{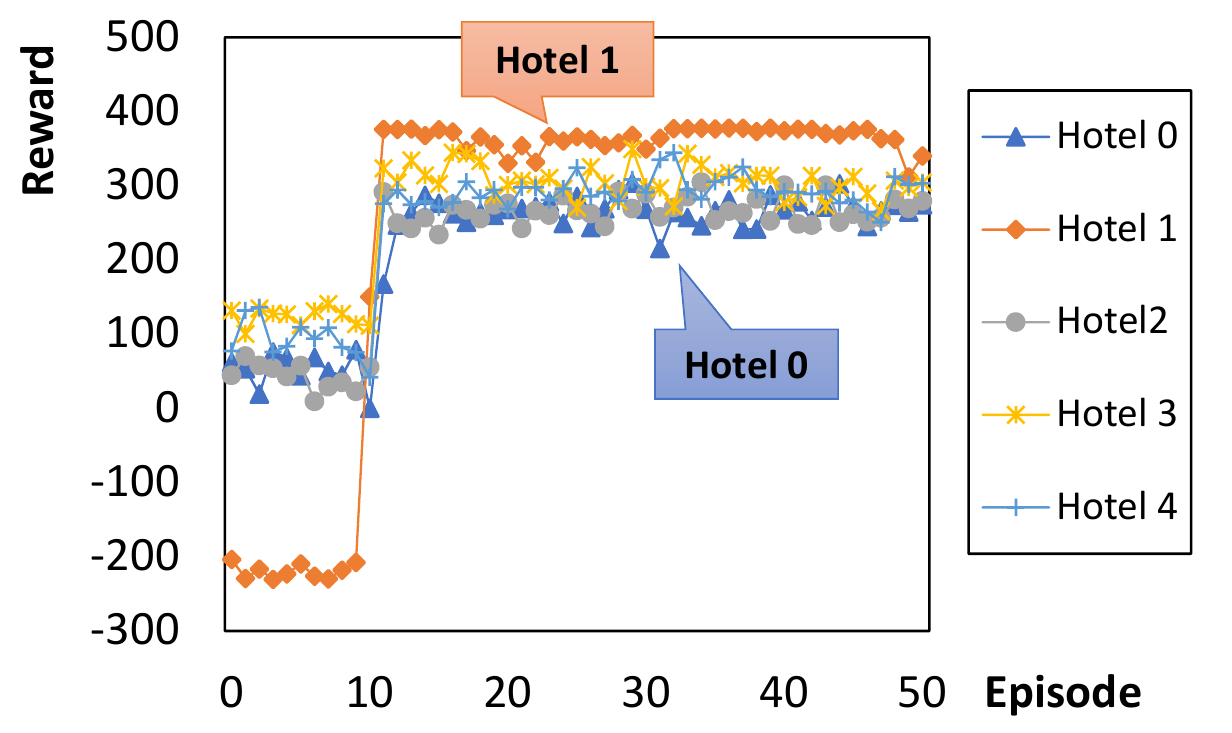}
		\caption{Hotel 1 is a CFT agent, others are DDPG agents}
		\label{fig:subfig:c}
	\end{subfigure}
		\caption{Comparison of different actors thinking with counterfactual actor-critic reinforcement learning on MSM with HOTEL dataset.}
		\label{fig:subfig} 
\end{figure*}
%
%

\subsection{Scalability}

We also compare the scalability for all related methods under MAWW in this section. To make a fair comparison, we set this environment with 2 same agents of each type (MADDPG, CFT, and CMPG) and all scalability experiments are completed on a workstation with E3 CPU, 64 GB RAM, and Nvidia P5000 GPU. Besides, we set the steps to explore in each episode to 100, the batch size for sampling to 100 and the exploration episode to 10 for all agents. For every CFT agent in this experiment, its scenario number $L=16$ and the parallel policy number $K=4$ (which is the same settings as in \uppercase\expandafter{\romannumeral4}.B). The result is shown in Figure 6. As it is shown in Figure 6, the computation efficiency of CFT is linear to the number of agents. We can observe that from Figure 6 (a), since the CMPG method needs to compute a normalized Q-value based on all previous actions in the replay buffer, it has the worst efficiency of all related methods. Furthermore, since MADDPG only uses a one-way agent to generate one exploration action and learns to update the action based on a single current Q-value from an ordinary critic, it is the most efficient method of all mentioned methods. Our CFT method uses a parallel structure to search several policy sub-spaces simultaneously, therefore, it is less efficient than MADDPG. However, it is still a more efficient method by comparing to CMPG. Figure 6 (b) shows the scalability of the CFT agent towards numbers of intent actions. We observe that the CFT agents are very efficient with the parallel policy number $K$ (from 2 to 10) and the computation time of CFT agents is linear to the intended action number. Therefore, CFT has the potential to be applied to large scale multi-agent reinforcement learning problems.

\begin{figure}[!t]
	\centering
	\begin{subfigure}[b]{3.2in}
		\includegraphics[width=\textwidth]{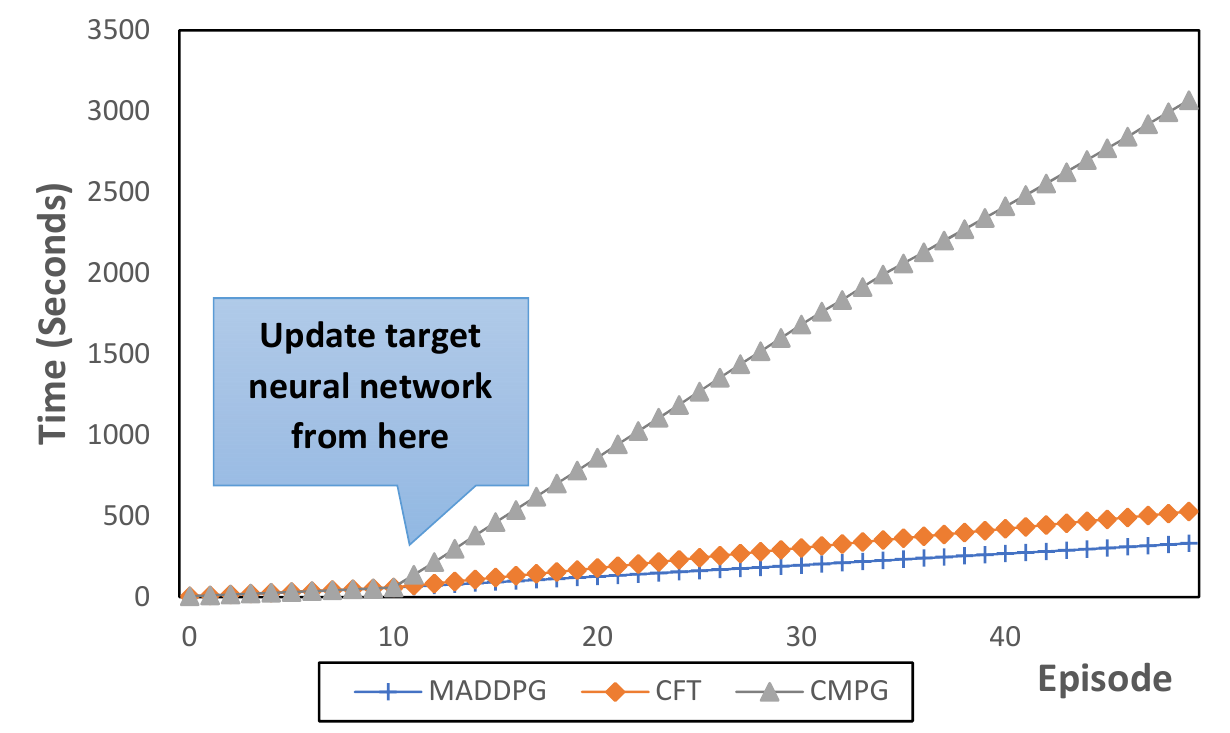}
		\caption{MADDPG v.s. CFT v.s. CMPG.}
		\label{fig:subfig:a}
	\end{subfigure}
	\begin{subfigure}[b]{3.2in}
		\includegraphics[width=\textwidth]{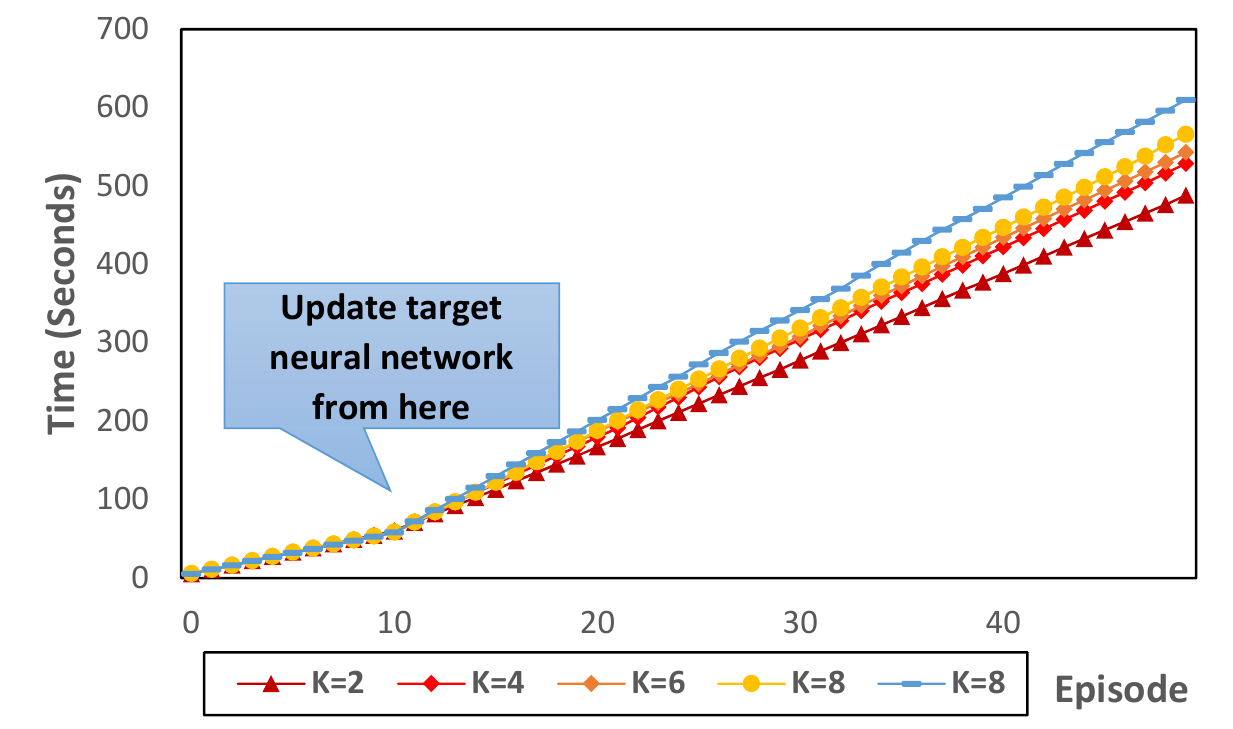}
		\caption{Comparison of CFTs with different intent policy number $K$.}
		\label{fig:subfig:b}
	\end{subfigure}
	\caption{Comparison of scalability for all methods.}
	\label{fig:subfig} 
\end{figure}

\section{Related Work}\label{sec_relatedwork}

Counterfactuals refer to the hypothetical states which are opposite to the facts \cite{pearl2009}. In studies about the complex systems which can hardly be accurately recreated, scientists usually use this idea to infer the consequences for unobserved conditions. e.g. medical scientists apply counterfactuals to discover the reasons to cause a certain disease \cite{HM2005}; psychologists use counterfactual thinking to predict future decisions of people in similar tasks \cite{Beldarrain2005}; historians infer the causal importance of some special events via counterfactual thinking \cite{SLevy2015}.

Reinforcement learning (RL) is an area of machine learning to train agents to take ideal actions in a dynamic environment in order to receive maximum accumulative rewards. Multi-agent deep reinforcement learning (MADRL) is the recent extension of RL to deal with the similar problems in high-dimensional environments \cite{DBLP:journals/corr/abs-1812-11794}. The hot topics for MADRL include learning to communicate between agents \cite{DBLP:conf/nips/FoersterAFW16}, exploring the competitive or cooperative behavior patterns for agents \cite{DBLP:conf/nips/LoweWTHAM17}, etc.

The main challenge to effectively train MADRL models is to explore as much as policy subspaces with limit observations. One reason for this problem is that it is difficult to completely recreate a high-dimensional multi-agent environment in which every agent behaves exactly the same as itself in history. Therefore, the observed action-reward tuples from the running environment are usually sparse and this sparse observation hinders the convergence rate for MADRL models. Counterfactual thinking shed a light on this issue by maximizing the utilization of observations to improve the learning efficiency. Concretely, to incorporate the counterfactual information into the process of reinforcement learning. Wolpert et al. \cite{doi:10.1142/9789812777263_0020} proposed the difference reward to revise the original rewards of an agent by the rewards under a default action during the simulation process. Jakob N. Foerster et al. \cite{DBLP:conf/aaai/FoersterFANW18} applies the average of all historical actions of an agent as the estimation for Q-values under counterfactual actions. This method use a regularized reward as the estimation for the real reward to compute the current Q-value for critics. All previous methods improve the performance of agents under multi-agent settings. However, they still do not directly address the problem to increase the efficiency for exploring the policy subspaces.

To enlarge the exploration coverage of policy subspaces for agents at each iteration, our method implements the counterfactual thinking by mimicking the human psychobiological process \cite{Beldarrain2005} with intent generating and evaluating with current experience. The experimental results show that this indeed improves the learning efficiency for an agent in the multi-agent environment and make it more competitive.

\section{Conclusion}\label{sec_conclusion}
In the multi-agent environment, it is difficult to completely recreate a historical moment for an environment since this needs to replay all actions for the related agents in the same order historically. Therefore, if an agent has choices at a special moment in a multi-agent environment, it challenges a lot to compute the accurate results for actions other than the practical chosen one. In order to estimate the possible returns for those non-chosen options, we propose the counterfactual thinking multi-agent deep reinforcement learning model (CFT). This model generates several intent actions which mimic the human psychological process and then learns the regrets for the non-chosen actions with its estimated Q-values at that moment simultaneously. The estimated Q-values and policies of an agent supervise each other during the training process to generate more effective policies. Since this framework can explore the policy subspace parallelly, CFT could converge to the optimal faster than other existing methods. We test CFT on standard multi-agent deep reinforcement learning platforms and real-world problems. The results show that CFT significantly improves the competitive ability of a specific agent by receiving more accumulative rewards than others in multi-agent environments. This also verifies that the counterfactual thinking mechanism is useful in training agent to solve the multi-agent deep reinforcement learning problems.

\section{Acknowledgements}
This work is supported by the National Natural Science Foundation of China (Grant No. 61602535,61503422), Program for Innovation Research in Central University of Finance and
Economics, Beijing Social Science Foundation (Grant No. 15JGC150), and the Foundation of State Key Laboratory of Cognitive Intelligence (Grant No. COGOSC-20190002), iFLYTEK, P.R. China. This work is also supported in part by NSF under grants III-1526499, III-1763325, III-1909323, SaTC-1930941, and CNS-1626432 .

\bibliographystyle{unsrt}
\bibliographystyle{IEEEtran}
\bibliography{ref}

\end{document}